# Annotating Electronic Medical Records for Question Answering

Preethi Raghavan, PhD[1], Siddharth Patwardhan, PhD[1],
Jennifer J. Liang, MD[1], Murthy V. Devarakonda, PhD[1]
[1]IBM T. J. Watson Research Center, Yorktown Heights, NY, USA

**Abstract**

*Our research is in the relatively unexplored area of question answering technologies for patient-specific questions over their electronic health records. A large dataset of human expert curated question and answer pairs is an important pre-requisite for developing, training and evaluating any question answering system that is powered by machine learning. In this paper, we describe a process for creating such a dataset of questions and answers. Our methodology is replicable, can be conducted by medical students as annotators, and results in high inter-annotator agreement (0.71 Cohen's κ). Over the course of 11 months, 11 medical students followed our annotation methodology, resulting in a question answering dataset of 5696 questions over 71 patient records, of which 1747 questions have corresponding answers generated by the medical students.*

**Introduction**

In recent years, automatic question answering (QA) systems have made big strides in addressing many different types of question answering problems. IBM's Watson[1] successfully challenged human grand champions on a televised quiz show of *Jeopardy!* Similarly, systems have also been built for answering factoid questions from structured resources[2,3,4,5], for answering questions about news stories[6] and even for handling 4$^{th}$ grade science questions[7]. In the medical domain, similar systems[8,9] have been designed to answer questions on medical knowledge using resources such as MEDLINE abstracts, PubMed Central full-text articles, eMedicine documents, etc.

Patient-specific QA from electronic medical records (EMRs), however, is a new and relatively unexplored problem in clinical natural language processing. The idea is for such a system to be available to physicians during patient visits, for instance, for easy access to information about the patient. A system that would automatically answer patient-specific questions by analyzing the patient's longitudinal medical records, including clinical narratives such as discharge summaries, progress notes, radiology reports as well as structured procedure, allergy and diagnoses lists, would be of immense value to medical professionals. Using such a system, a physician could get quick answers to questions such as

> Has this patient had a surgical procedure recently?

> What medications is the patient taking to control his hypertension?

> Does she smoke?

Answers to these questions are facts about the patient that would be retrieved or inferred by the system from the patient's EMR.

We refer to this kind of QA as patient-specific QA, and, although this type of QA is also in the medical domain, it differs from previous QA research initiatives in at least two ways. First, the answers to these questions are not well known facts about the world, but are very specific facts about a specific patient. This typically requires searching for a particular word, phrase or passage in the patient EMR, with the evidence for that fact being localized to one small part of the EMR. Compare this with the factoid/trivia QA, where the fact, as well as the evidence for that fact can be found in many different parts of the underlying knowledge resources. The second major difference for patient specific QA is the lack of question-answer datasets required to train, develop and evaluate such systems.

Most QA systems rely on machine learning technology that learn from training examples of question-answer pairs. To successfully train a machine-learning model for QA, we require the training questions to be realistic, and representative of questions that would be asked by a user of the system. For patient-specific QA, this requirement makes it especially challenging and expensive to obtain a set of such question-answer pairs. It is no easy task to automatically gather questions that a physician would ask about her patients during the course of her day. Furthermore, the correct answers to these questions are words, phrases or passages in the patient's EMR, and must be manually identified by a medical expert – again, no easy feat.

In this paper, we describe the creation of a dataset of manually curated question-answer pairs for patient-specific QA. We developed a process for generating questions that are likely to be asked by a physician in a real clinical setting, and we also defined a process for annotating corresponding answers from both structured and unstructured parts of the patient's EMR. The main contributions of this paper are as follows:

(1) We outline a replicable annotation process, with minimal annotator bias, for generating questions similar to what a physician may want to ask about a patient.

(2) We also outline a replicable annotation process for marking answers to the generated questions in the EMR, and then discuss the inter-annotator reliability and factors leading to annotator disagreement.

(3) We present an analysis of the dataset created, as well as some limitations of our methodology.

(4) Our methodology is replicable, can be conducted by medical students as annotators, and results in high inter-annotator agreement (0.71 Cohen's κ).

(5) Eleven medical students created a question answering dataset of 5696 questions over 71 patient records following our annotation methodology. Of these questions, 1747 have corresponding answers generated by the same medical students. This data was created over a period of 11 months, but the work was conducted intermittently over this period.

In the following sections, we cover the related work from literature, a description of our methods, and our observations on the generated data set.

**Related Work**

There have been multiple studies that generate resources and build systems for biomedical question answering.[8,9,10,11] The common question types, semantic models of questions, as well as resources and systems available for biomedical question answering have been well-documented.[8,9] Current state-of-the-art biomedical QA systems primarily deal with medical questions that relate to patient-care but are not patient-specific.[8,9] Thus, the resources used to answer biomedical questions are typically MEDLINE abstracts, PubMed Central full-text articles, eMedicine documents, clinical guidelines and Wikipedia articles. In contrast, patient-specific questions are typically answered from the structured and unstructured resources in the EMR.

Ely et al.[11] developed a taxonomy of medical questions collected from primary care doctors in a clinical setting. 64 generic question types were used to classify 1396 clinical questions. The three common generic types were "What is the drug of choice for condition x?" "What is the cause of symptom x?" and "What test is indicated in situation x?" Some of the questions in this dataset have a patient-specific nature, for example "Could this patient have condition x?" and "Could this patient have condition x?" While this question set is used by multiple medical QA systems, these systems do not address the problem of answering patient-specific questions and hence do not gather answers to the patient-specific questions.

ClinicalQuestions or Clinques (http://clinques.nlm.nih.gov) is a question repository of 4654 questions maintained by the NLM that has questions collected from healthcare providers in clinical settings across the US.[11,12] Most biomedical QA systems either use the Clinques repository[13,14] of questions or create their own question set for manual evaluation by physicians. The MiPACQ[14] system narrowed the questions to those likely to be answered by the MEDPEDIA and excluded questions that required patient-specific knowledge. A few other datasets have also been created for biomedical QA through different question answering challenges. The BioASQ challenge[15] uses benchmark datasets of about 500 questions and gold-standard answers constructed by a team of biomedical experts from around Europe. The answers consist of relevant concepts from designated terminologies, articles, snippets, RDF triples from designated ontologies, exact answers and paragraph-sized summaries. Denmer-Fushman et al.[16] use the TREC 2007 genomics dataset to identify answers in the form of semantic relations between biomedical entities in questions and potential answer types.

Community-shared annotated datasets for clinical NLP tasks have been developed as part of the i2b2 challenge[17], CLEF[18] and SemEval[19]. However, there is no systematically developed resource of questions and answers available to develop a robust patient-specific question answering system.

**Methods: Annotation Process**

The objective of the annotation task was to generate questions that physicians would ask of the EMR in a clinical setting. The answers to these questions, if found in the EMR, maybe be one or more phrases, passages. Annotated data is used to build, train and evaluate models for the EMR QA task. This implicitly assumes that the annotations represent the truth. However, this assumption can be violated in two ways: either because the annotators exhibit a certain bias, or because what is considered as truth may be subjective. We propose an annotation process for question answering on EMRs developed using de-identified medical records available to us through a research collaboration with Cleveland Clinic, that is aware of these problems and tries to address them to a large extent. The annotation task requires annotators with medical domain expertise. The annotators used in our EMR QA annotation process are medical students in their third or fourth year of training. The annotation process is divided into two stages i.e. question generation and answer generation.

*Question Generation*

The question generation process is designed to generate questions that (1) minimize bias in question language, (2) are representative of what physicians would ask of a patient's EMR in a clinical setting, and (3) cover most common topics of interest to physicians.

One important concern is that the language in the EMR, when formulating a question, should not bias the annotators. This would lead to questions being worded in a manner that makes it easy to find the answer. For instance, if an annotator reads a clinical note that mentions "the most recent colonoscopy was done in 2005", this may prompt them to formulating the question "when was the most recent colonoscopy done?" Machine learning models trained to predict passage or answer relevance for a particular question learn predictive relationships based on patterns found in the training set. Given such biased data, the trained models are likely to perform poorly on questions generated in a real setting. We try to minimize this bias by not allowing the annotators to read through the entire EMR during question generation. Instead, they are presented with the following starting points:

(1) The Watson Electronic Medical Record Analysis (EMRA) summarization user interface (Figure 1). This is a dashboard that provides a view of automatically generated problem lists, structured medications, labs, procedures, and a view of clinical notes over time.[20,21]

(2) The latest clinical note or any well formed clinical note (that has a reason for visit, assessment and plan).

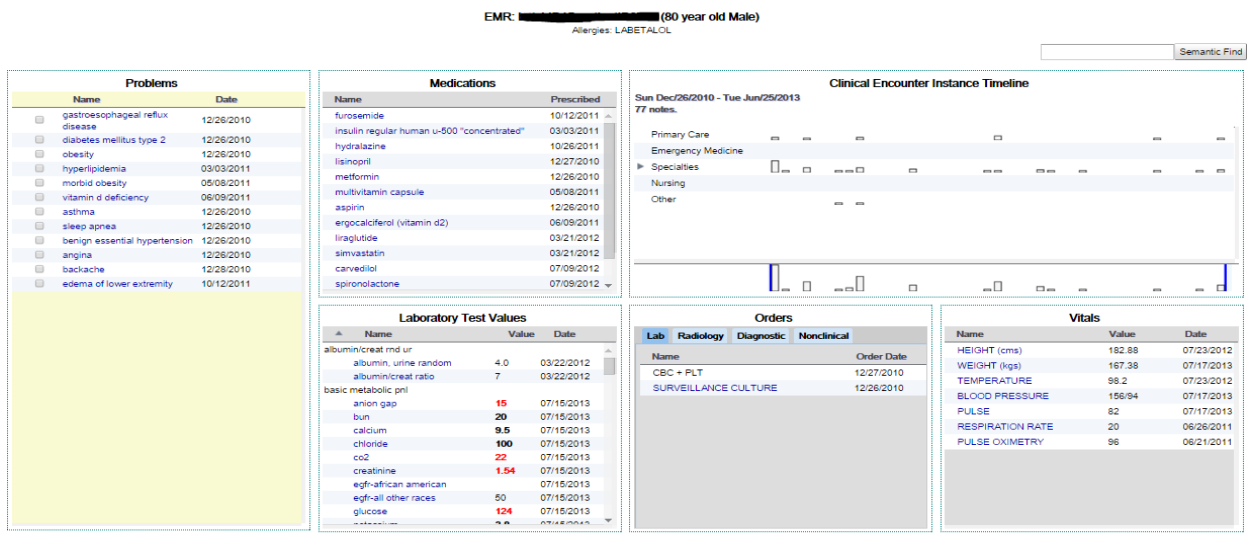

**Figure 1.** Screenshot of Watson EMRA summarization UI.

During review of Watson EMRA summarization, the annotators ask questions that get trigged based on the information they see i.e. problem list, medications, vitals etc. For the latest clinical note, the annotators ask questions that get trigged based on what they read. If the note presented to them does not trigger many questions, they may pick another well-formed note. In both cases, the questions need not be about the medical entities present on the summarization UI or in the note reviewed, but any question that occurs to the annotator based on what they read review. These starting points for generating questions were meant to simulate what physicians might ask at the point-of-care, where they may review a cover page or the patient's most recent progress note to get a quick summary of the patient prior to the patient encounter (Figure 2). Thus, the annotators are given the following instructions:

"Imagine yourself as a clinician about to see a patient. As preparation for the upcoming visit, you review the patient's information by viewing either Watson EMRA Summarization or the patient's last informative note. What are some questions you would like to ask on the patient's EMR to help you in preparing for this visit?"

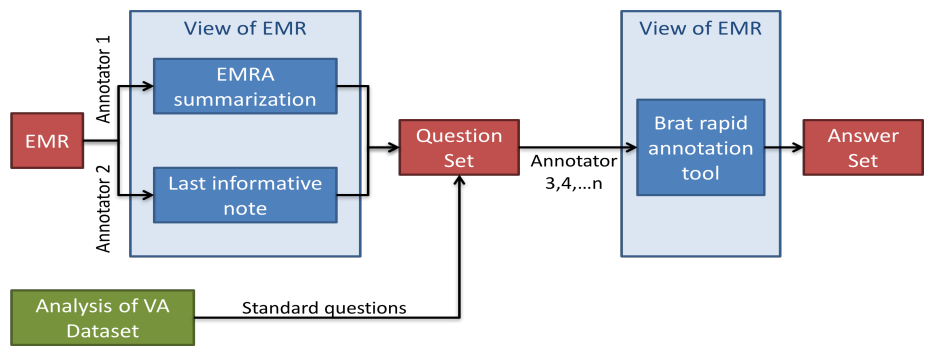

**Figure 2.** Overview of annotation process for question and answer generation.

Generated questions were captured using a spreadsheet template provided to the annotators that includes the following fields: EMR ID, question ID, generated question, the specific context that triggered the question (e.g. "Blood pressure 149/93") and the origin or source of that context (i.e. panel within summarization or note ID of last informative note reviewed).

To ensure that the questions generated would be representative of the topics most of interest to physicians, we (1) provide to the annotators a list of the general types of questions asked by physicians on EMRs, and (2) include standard questions on common topics that were asked on every patient EMR (Table 1).

**Table 1.** List of common question topics and their counts within the VA Dataset of 976 questions, along with the standard questions added to our dataset to be asked on all patient EMRs.

| Common Topic | Count in VA Dataset | Standard Question |
|---|---|---|
| Colonoscopy | 46 | When was the last colonoscopy? |
| | | What were the results of the last colonoscopy? |
| Echocardiogram | 25 | Has the patient had an echocardiogram? |
| | | What are the results of the last echocardiogram? |
| Pap smear | 13 | When was the last pap smear? |
| Mammogram | 12 | When was the last mammogram? |
| Antihypertensives | 12 | What antihypertensives has the patient tried? |
| Antidepressants | 11 | What antidepressants has the patient been tried on? |
| Substance Abuse | 15 | Does the patient have history of substance abuse? |
| Smoking | 8 | When did the patient quit smoking? |

We extracted the lists of question types and standard questions from a set of questions asked by physicians at the Veterans Hospital Administration (VA) on patient medical records. This data was provided to us through a research collaboration agreement with VA, and was used as a starting point to understand the different types of questions that are asked by physicians in a real clinical setting. We have approximately 976 questions from the VA.

Questions are required to be complete sentences, self-contained and potentially answerable on EMR content alone. Note that questions are required to only be potentially answerable; this means that the answer to these questions may or may not be found in the EMR, a quality that reflects those questions asked in real-world clinical settings.

Given this annotation process, each annotator is able to generate 10 to 20 questions for a single EMR in one hour. We have generated a total of 5696 questions over 71 patient records so far using this process, 2492 of which have been analyzed and the results discussed later in this paper.

*Answer Generation*

Given a set of questions paired with EMRs generated in the question generation phase, the annotators then annotate answers to those questions within the context of each clinical note (Figure 2). We describe here our final annotation process converged on after multiple pilot annotation tasks. This process generates a consistent and complete answer set for each question with good inter-annotator agreement (IAA), and records the context of the marked answer within the scope of the full note. For now, this process has only been applied for generating answers on unstructured notes. Ongoing work involves extending this process to generate answers on structured data within the EMR.

Annotations for answer generation were completed using the brat rapid annotation tool (brat), which supports text span annotations within documents as named entities.[22] To adapt this tool for our purposes, for each EMR, all clinical notes were ordered chronologically and loaded into brat, and annotators were provided with a list of questions to be answered with corresponding Question IDs. The Question IDs are used in brat to represent questions to be answered on that same EMR. Questions are answered by a different annotator than the ones who generated the questions to eliminate any bias from previous review of the EMR during the question generation process.

Figure 3 shows the brat user interface used by annotators to record answers to questions. At the time when this screenshot was captured, the annotator has already marked two spans of text "diabetes for the last 10 years" and "diabetic for the last 10 years" as answers Question ID Q05 ("How long has he had diabetes?"), and is in the process of annotating a new span "home medications include insulin" as the answer to Question ID Q06 ("Has the patient ever taken insulin for their diabetes?").

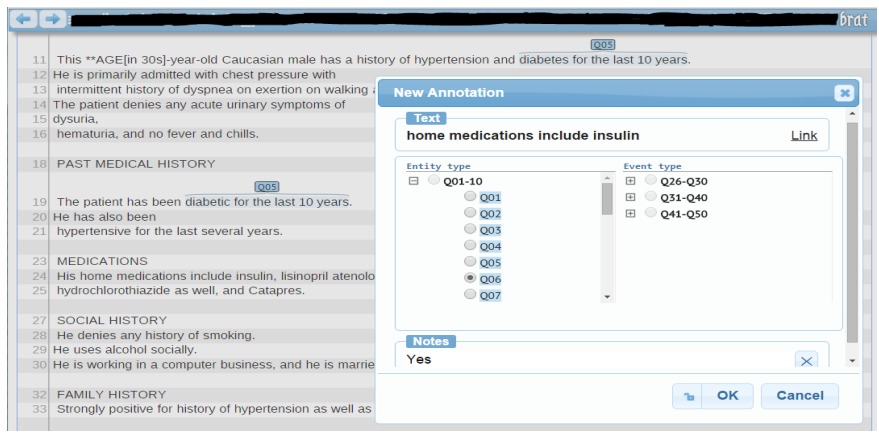

**Figure 3.** Brat user interface used to annotate answers within a clinical note.

With this setup, an annotator performs a note-by-note review of the entire EMR and annotates any text spans that answer a question with the corresponding entity. A note-by-note review ensures completeness of the annotations so that no answers are missed. If the annotated span does not reflect the final answer to a question and requires some inference to come to the answer, a free-text Notes field allows the annotator to enter the final answer to the question. This is most frequently found in yes/no type questions, but may apply to other types of questions as well.

As questions with temporal qualifiers are difficult to answer before having reviewed the entire EMR, annotators were asked to disregard temporal qualifiers while annotating answers during this note-by-note review within brat. So "What are the results of his colonoscopy?" and "What are the results of his last colonoscopy?" would have the same spans annotated as answers. For any questions with temporal qualifiers, a separate pass on the spans annotated as answers would be required to determine the final answer.

**Results and Discussion**

*Question Generation: Analysis of dataset*

Out of the total 5696 questions generated, for a subset of 2492 questions, we studied the distribution of questions generated using different contextual triggers i.e. various medical concepts that triggered that specific question in the last informative note, standard questions or the summarization UI. The most frequent context triggers along with their counts are shown is shown in Table 2. The summarization contextual triggers like problems, orders, medications etc. are displayed on different panels of the summarization UI shown in Figure 1.

**Table 2.** Examples of generated questions along with the corresponding context trigger.

| Context - Source | Count | Example Context | Example Questions |
|---|---|---|---|
| Last informative note | 914 | Past Medical History of - 11-2206 (vasculopothy/OCP induced) | Does the patient have a family history of blood disorder? |
| Standard questions | 475 | | What antihypertensives has the patient tried? |
| Summarization-Problems | 680 | Hyperlipidemia | When was the patient's last lipid panel performed? |
| Summarization-Orders | 111 | dilated fundus exam | Has there been a systematic increase in this patient's urine microalbuminuria? |
| Summarization-Medications | 76 | baclofen | How long has the patient been on baclofen? |
| Summarization-Demographics | 58 | 64 year old Female | When was the patient's last mammogram? |
| Summarization-Labs | 36 | CA 125 bld | Why was a cancer antigen, ca-125 test performed? |
| Summarization-Allergies | 23 | epinephrine | What kind of reaction does the patient have to epinephrine? |

We also analyzed the types of questions from the subset of 2492 questions (Figure 4). While there could be many more question types, we have identified a handful of types that help us make sense of the generated questions from a question analysis and answer generation perspective. A question may occur across multiple types.

**Temporal:** This question type includes "when" questions, questions with temporal qualifiers like "current", "previous", "next", "last" (and their inflections) and duration-based questions. Since patient records are longitudinal in nature, the large number of temporal questions is not surprising. In analyzing this type of question, it is important to understand dependencies between temporal keywords and medical concepts in the question. For instance, in the question, "Has the patient had a previous MI?" we should identify previous as a temporal qualifier applied to the medical concept MI (myocardial infarction).

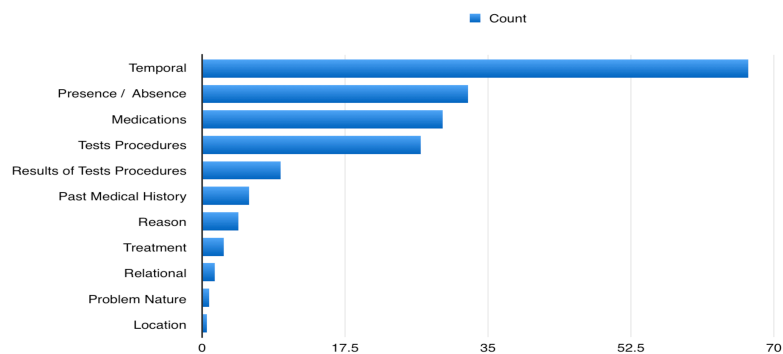

**Figure 4.** Question type distribution

**Presence/Absence:** Questions in this type need confirmation of the presence or absence of a problem, test or procedure. While these questions may appear to be like the factoid "yes/no" style questions, the expected answer for these questions goes beyond a simple "yes" or "no". When a physician asks "Did the patient have a colonoscopy?" not only may he want to ascertain that the procedure happened, but also find out about when and where it happened, along with the outcome of the procedure.

**Medications, Tests, Procedures:** Medication questions typically ask about the medications taken by the patient for a particular problem (both currently and in the past), allergic reactions to specific medications, when a medication was started or stopped. A few of them also ask for the "reason" why the patient is on a particular medication. These questions also get classified under the "reason" question type. Question analysis on medication type questions like "What *medications* has the patient tried for *hypertension in the past*?" needs to identify the medical concepts (*hypertension*) on which the medication relation (*mediations for*) is to be applied. We also need to account for the temporal aspect in the question by applying the temporal relation (*in the past*) to the results of the medication relation. In case of labs and procedures, the questions tend to ask if a particular test/procedure was done, when and why it was done, and the results of the test or procedure. Similar to the medications type, the questions in this type could have temporal or reason type characteristics.

Other more infrequent question types (some of which overlap with many of the previous types like temporal, medications, tests/procedures) include "**Past Medical History**" that consists of questions asking about problems that have occurred in the patient's history and "**Problem Nature**" that discusses the severity, recurrence and status of the problem. Another type "**Location**" has questions about the location of a problem in the patient's body. Instances of the "**Reason**" question type were seen in the medications, tests and procedures type. Finally, the "**Relational**" question type has multiple medical concepts that have to be considered in relationship with each other to answer the question correctly. For instance, in the question, "Is the patient's pain due to generalized osteoarthritis well-controlled with medication?" we need to correlate the pain caused by the problem of generalized osteoarthritis with the its corresponding medication. This category of questions, though infrequent in our dataset, can be of high value to the physician because even with several clicks in an EHR system, the physician may be unable to make this correlation without pulling out multiple pieces of information and analyzing them in conjunction with each other.

Another characteristic of the questions generated in the annotation process is that they are mostly fact-based i.e. the answer is understood to exist somewhere in the EMR (although this may not always be true), and the questions seek to unravel those answers buried in the unstructured and structured resources in the EMR. An important category of questions is based on what is not observed in the EMR. For example, when a physician observes a patient who has a myocardial infarction but is not on any ACE inhibitors, he may ask "Why is the patient not on ACE inhibitors?" The answer may be that the patient is allergic to that category of medications or has experienced some adverse effect in the past due to ACE inhibitors, or the answer may not be apparent from any related information found in the EMR. Generating more of these types of questions may require some tweaks to our annotation process such as encouraging annotators to also think about patient questions based on what is missing from the EMR. Incorporating this aspect into our annotation is something we intend to do in future iterations of our annotation task.

*Answer Generation: Development of annotation process and inter-annotator agreement*

To arrive at the answer generation process described previously, a total of five pilots were conducted to hone various aspects of the annotation process. For each pilot, 3 EMRs with approximately 40 questions each was double annotated, then reviewed for reconciliation, and a post-pilot discussion was held with researchers and annotators together to explore issues that arose during the pilot. Inter-annotator agreement was also computed at the end of each pilot to enable us to develop a more robust annotation process that minimizes annotator-bias and enforces consistency in answering questions.

Our initial pilot mimicked the natural process of finding an answer to a question, where given a question, an individual browses and/or searches through the EMR to find the answer. With this as our starting point, we found in our earlier pilots that there were essentially two types of disagreements: (1) actual disagreements due to differences in what is considered an answer and (2) apparent disagreements that were simply answers overlooked by one of the two annotators due to differences in their method of searching and navigating through EMRs.

Two examples of the first type of disagreement are listed below:

(1) For questions asking whether a concept was present in the patient (e.g. "Does the patient use a CPAP machine?"), one annotator considered the absence of a concept to indicate negation – and therefore an answer, while another annotator did not – instead considering this as a question with no answer.

(2) For questions asking about the existence of a disease in the patient, one annotator considered the disease to be present if the findings needed to meet the criteria for the diagnosis were available and consistent with the diagnosis (e.g. considering two elevated blood pressure readings taken weeks apart to indicate a diagnosis of hypertension), while another annotator considered only those diagnoses that were explicitly stated within the medical record as answers.

Subsequent pilots implemented guideline changes to clearly define what constitutes an answer, and annotation process and tooling changes to minimize overlooked answers and reduce annotation time per EMR. Table 3 summarizes the lessons learned and adjustments made with each pilot, along with IAA calculated at each stage. The reported IAA is pre reconciliation and at the answer note ID level.

**Table 3.** Overview of pilots conducted for answer generation. Cohen's Kappa was used to measure inter-annotator agreement.

| Pilot | Tooling | Changes implemented in this pilot | Lessons learned from this pilot | Kappa |
|---|---|---|---|---|
| Pilot 1 | Spreadsheet | *n/a* | Confusion regarding what constitutes an answer. Bias observed when annotator who generated the question also answered the question. | 0.36 |
| Pilot 2 | Spreadsheet | Guidelines revised to clarify what should be considered as an answer. Annotators assigned so that the person answering the question is different from the person who generated the question. | Observed two types of disagreement: (1) true disagreements due to differences in judgment, and (2) apparent disagreements due to answers missed by annotator during annotation process. | 0.54 |
| Pilot 3 | Spreadsheet | Implemented process changes to reduce "false" disagreements. | Slight improvement in IAA with process changes, but still observing missed answers. | 0.63 |
| Pilot 4 | IBM Human Annotation Tool | Switched to note-by-note review to ensure completeness of annotations. Moved to using IBM Human Annotation Tool for annotation to facilitate note-by-note review. | A note-by-note review increased IAA and reduced annotation time by more than half. Presence of annotation errors due to difficulties in (1) viewing annotations when a span of text is annotated more than once, and (2) reviewing annotations when correction is needed. | 0.71 |
| Pilot 5 | Brat rapid annotation tool[22] | Moved to using brat, which allows for easier visualization of multiple overlapping annotations, and has a search function to locate specific text or previously annotated spans. Converted to single annotator annotations because relatively good IAA with note-by-note review. | General positive feedback from annotators. Completion of pilot process. | *n/a* |

Generated answers can be broadly categorized by (1) if the answer is expressed as a fact from the patient record or requires one or more paragraphs of explanation, and (2) if the answer is explicitly stated or requires inference.

Answers that consist of facts usually correspond to **Location** or **Presence/Absence** questions, as well as a subset of **Temporal** questions that ask for a specific time or date (for e.g., "when" questions); while answers that involve paragraphs of explanation are mostly **Problem Nature** or **Reason** questions (e.g. "why" questions, "what was the cause of x?"). Questions on **Medications**, **Tests**, and **Procedures** vary depending on the specific information requested; questions on whether a test was done usually have answers expressed as facts, while questions on why a test was done or the results of a test may require a paragraph as an answer.

Answers that require inference can be further separated into common-sense logical inferences that are made using general knowledge, and those inferences that require domain-specific knowledge. For example, the question "Is his hoarseness caused by the thyroid surgery?" can be answered as "No" if there is documentation of hoarseness in the patient prior to thyroid surgery. This is an inference that can be made without domain-specific knowledge. This is in contrast with the following example. For the question "Does the patient have diabetes?", documentation of the patient being prescribed the medication metformin would be annotated as an answer, considering that this medication is specifically used for treatment of diabetes so would only be prescribed if the patient has this problem. This is an inference that requires medical knowledge.

Examples of different types of answers are shown in Table 4.

**Table 4.** Sample question-answer pairs.

| Question | Answer Passage | Answer | Answer Type | Inference Required |
|---|---|---|---|---|
| Why does the patient take trazodone? | 5. 780.52 Insomnia<br><br>- TRAZODONE 50 MG TABLET | Insomnia | Fact | None |
| Does the patient currently smoke? | Tobacco Use: Quit Packs/Day: .5 Years: 1.5 | No | Fact | Logical |
| Has the patient ever been pregnant? | This is a 40 year old G8P3325 who presents for her annual gynecologic exam. | Yes | Fact | Medical |
| What were the results of the patient's most recent echocardiogram? | LABORATORY DATA PRIOR TO ADMISSION Included normal echocardiogram with mild pulmonic valve insufficiency and moderate left ventricular hypertrophy. | normal echocardiogram with mild pulmonic valve insufficiency and moderate left ventricular hypertrophy | Paragraph | None |

Differentiating between these types of answers becomes necessary as we move ahead with developing a question-answering system on EMRs and evaluating our performance for these various types of question-answer pairs.

**Limitations**

We summarize some of the limitations of our annotation process and the reasons behind them.
 (1) **Questions are required to be complete sentences**. A question expressed as "When was the last colonoscopy?" tells us that the physician specifically wants to know when the last colonoscopy was performed, whereas an incomplete query like "last colonoscopy" could mean that the physician may want to know "when", "where" or the "results" of the procedure. However, in a real clinical setting, a physician may prefer entering short phrases or keywords (which may be more appropriate for traditional keyword-based search) instead of complete sentences.
 (2) **Questions are required to be self-contained**. The "context" information used to generate questions is not used for answer generation.
 (3) **Questions are potentially answerable on EMR content alone**. Some questions like "Does the patient have any neurological symptoms?" may require information from both the EMR as well as external medical resources to provide a complete answer. We do not generate such questions.
 (4) **Brat allows us to only view a single clinical note at a time**. Thus, it does not support the annotation of cross-document relations and restricts the marking of related answer spans to a single note. While there may be ways to work around this, annotating relations across a longitudinal EMR is a very tedious task.
 (5) **Simulated clinical setting**. Medical student annotators simulate the process of a physician asking patient-specific questions in a real setting where he is treating a patient. In this simulated setting, the students may lack the relevant context to generate certain complex questions.

**Conclusions**

We have defined a replicable methodology for generating a question and answer dataset for the relatively unexplored problem of patient-specific QA over EMRs. The annotation task was conducted using medical student annotators. The question generation process minimizes annotator bias, while simulating a physician asking questions in a real clinical setting. The answer generation process identifies appropriate phrases, passages or notes based on the question type and how the answer is expressed in the EMR. Using this methodology, we generate a dataset of 5969 questions and 1747 answers. Our results show how different types of questions and answers may be generated with high inter-annotator agreement. Our analysis of the QA dataset provides insights into tackling the patient-specific QA task. We plan to release our question set, along with question type analysis, to encourage community participation in question analysis for patient-specific QA.